\documentclass[
twocolumn,
]{ceurart}

\usepackage{listings}
\begin{document}

\copyrightyear{2023}
\copyrightclause{Copyright © 2023 for this paper by its authors. Use permitted under Creative Commons License Attribution 4.0 International (CC BY 4.0).}

\conference{AISafety-SafeRL 2023 Workshop (IJCAI), August 19–21, 2023, Macao, SAR, China}


\title{Empirical Optimal Risk to Quantify Model Trustworthiness for Failure Detection}


\author[1]{Shuang Ao}[
orcid=0000-0003-2648-3082,
email=shuang.ao@open.ac.uk,
url=https://github.com/AoShuang92,
]
\cormark[1]


\author[2]{Stefan Rueger}[
email=stefan.rueger@open.ac.uk,
url=https://kmi.open.ac.uk/people/member/stefan-rueger,
]

\author[3]{Advaith Siddharthan}[
orcid=0000-0003-0796-8826,
email=advaith.siddharthan@open.ac.uk,
url=https://people.kmi.open.ac.uk/advaith-siddharthan/,
]

\address[1,2,3]{Knowledge Media Institute, The Open University, Walton Hall, Kents Hill, Milton Keynes MK7 6AA, UK}


\cortext[1]{Corresponding author.}

\begin{abstract}
 Failure detection (FD) in AI systems is a crucial safeguard for the deployment for safety-critical tasks. The common evaluation method of FD performance is the Risk-coverage (RC) curve, which reveals the trade-off between the data coverage rate and the performance on accepted data. One common way to quantify the RC curve by calculating the area under the RC curve. However, this metric does not inform on how suited any method is for FD, or what the optimal coverage rate should be. As FD aims to achieve higher performance with fewer data discarded, evaluating with partial coverage excluding the most uncertain samples is more intuitive and meaningful than full coverage. In addition, there is an optimal point in the coverage where the model could achieve ideal performance theoretically. We propose the Excess Area Under the Optimal RC Curve (E-AUoptRC), with the area in coverage from the optimal point to the full coverage. Further, the model performance at this optimal point can represent both model learning ability and calibration. We propose it as the Trust Index (TI), a complementary evaluation metric to the overall model accuracy. We report extensive experiments on three benchmark image datasets with ten variants of transformer and CNN models. Our results show that our proposed methods can better reflect the model trustworthiness than existing evaluation metrics. We further observe that the model with high overall accuracy does not always yield the high TI, which indicates the necessity of the proposed Trust Index as a complementary metric to the model overall accuracy. The code are available at \url{https://github.com/AoShuang92/optimal_risk}.

\end{abstract}


\maketitle

\section{Introduction}
\label{sec:intro}

The deployment of deep neural networks (DNNs) in safety-critical applications such as autonomous driving~\cite{atakishiyev2021explainable} and medical diagnosing~\cite{raghu2019direct,dusenberry2020analyzing} requires high trustworthiness and reliability, as mistakes can be expensive and raise serious concerns. To reduce mispredictions, a model should be equipped with a safeguard for automatic failure detection~\cite{hendrycks2017baseline, corbiere2019addressing, band2021benchmarking} or a reject option~\cite{hendrickx2021machine}, where samples with high uncertainty or low confidence can be discarded or sent to an expert or the third system. Specifically, failure detection (FD) determines the portion of coverage over the entire dataset deemed to be safe predictions and discards data using a threshold on model confidence or uncertainty. If the confidence or uncertainty is below or above the threshold, the model rejects samples and defers them to human experts or third systems to re-evaluate. Otherwise, the model considers these samples in a coverage range for safe and trusted prediction. FD is beneficial for gaining higher trust from users and for time and cost savings by only requiring human interventions for a small percentage of data.

One of the criteria for FD is for the model to achieve better performance with fewer instances removed; hence the evaluation is about the trade-off between the coverage of data and model accuracy or risk (error). Popular visualisation methods of FD performance such as risk-coverage (RC) curve~\cite{el2010foundations} and accuracy-rejection curves (ARCs)~\cite{ferri2004cautious, nadeem2009accuracy} plot model risk or accuracy against coverage of data. However, the quantification of FD performance is a less explored domain. Recent studies attempt to quantify FD by using the area under the RC-curve (AURC)~\cite{ding2020revisiting} and the area under the ARCs~\cite{nadeem2009accuracy}. Nevertheless, both methods include the full coverage of data, ignoring the selection of thresholds and the FD performance under and above thresholds.

\begin{figure*}[!h]
\centerline{\includegraphics[width=0.95\textwidth]{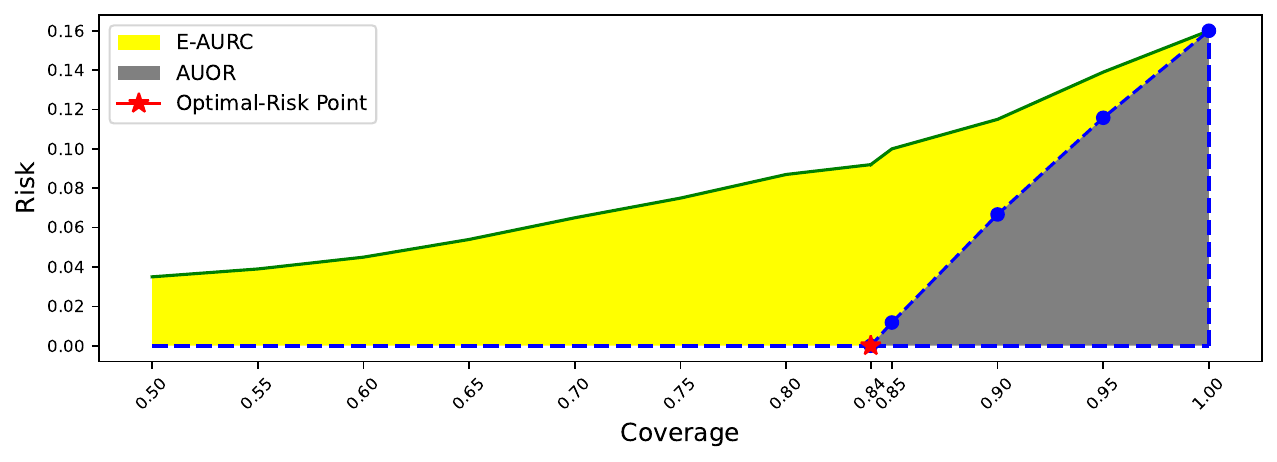}}
\caption{Risk-coverage curve for the ImageNet dataset with SwinTran model. The entire AURC is the yellow plus grey area, with the E-AURC shown as yellow area and the area under the optimal risk (AUOR) as the optimal-risk area. The optimal-risk point is at the coverage of model accuracy (in this case, 0.84). }
\label{fig:aurc}
\end{figure*}

Theoretically, a perfectly calibrated model should achieve the ideal performance (i.e., accuracy of 1) after removing the most uncertain samples in numbers equal to the error percentage. In other words, the perfect performance takes place hypothetically by covering the portion of samples equivalent to model accuracy. Therefore, the model risk is supposed to be 0 at this very coverage point, which is denoted as the optimal point in work on uncertainty estimation~\cite{geifmanbias} as shown in Figure~\ref{fig:aurc}. A perfectly calibrated model should not contain any risk before the optimal point, whereas the risk increases monotonically until the model error after the optimal point. This risk is naturally inherited from the model as DNNs cannot obtain the perfect performance in practice, thus, should perhaps be discounted in FD evaluations. Based on this hypothesis, Geifman et.al~\cite{geifmanbias} exclude the area under the optimal risk (grey part in Figure~\ref{fig:aurc}) for the AURC and propose the metric of Excess-AURC (E-AURC) (yellow part in Figure~\ref{fig:aurc}). However, this still evaluates FD based on the whole dataset even though some data are supposed to be safe and trusted predictions.

As the percentage of rejected samples is generally customised during deployment of a model, there is a lack of common ground for a fair comparison of failure detection among models with varying accuracies. In addition, most of the existing evaluation metrics (i.e., AURC, E-AURC) measure the entire area under the curve, which cannot reveal the FD performance for a specific coverage. For example, the performance of a model at very low coverage is not of interest to real applications. To address the above issues, we propose the Excess area under the optimal RC curve (E-AUoptRC) as an alternative metric for failure detection that considers the risk in the range from the optimal point to the full coverage (shown as pink area in Figure~\ref{fig:divaurc}). We emphasise this area for reasons as follows: (1) with a perfectly calibrated model, samples falling into the coverage from 0 to optimal point (yellow area in Figure~\ref{fig:divaurc}) are already highly trusted ones; (2) we argue that it is more important to compare models in the region that errors are made, for instance, samples in the E-AUoptRC include the high uncertainty ones, and the corresponding risk here should be primarily utilised to determine the trustworthiness of the model. (3) Furthermore, with our precise method of FD quantification, a model with lower accuracy may yield higher trustworthiness and vice versa, capturing the intuition that a model with higher accuracy may not be the most trusted one. Finally, we propose a Trust Index (TI) as a novel evaluation metric, which measures the accuracy of the model at the optimal point, mimics the behaviour of E-AUoptRC, and is easier to compute. The Trust Index combines the performance and calibration of the model into a single metric. A higher TI suggests better model performance and calibration and higher trust and reliability of the model predictions. 

Our contributions and findings are summarized as below: 

\begin{enumerate}

\item We propose the E-AUoptRC to quantify the RC curve with the coverage from the optimal point to the full coverage. 
\item We propose Trust Index as an evaluation metric. 
\item With extensive experiments and observations we find that: (i) a model with higher AURC or E-AURC can obtain lower E-AUoptRC ; (ii) A model with a high overall accuracy does not necessarily yield higher Trust Index; (iii) Our proposed methods can better evaluate failure detection for model trustworthiness. 

\end{enumerate}

\section{Related Work}
\label{sec:relate}

\subsection{Failure Detection}

In the deployment of safety-critical scenarios, DNNs tend to fail silently by providing high-confidence in woefully incorrect predictions, which makes the uncertainty estimation a great concern to AI safety~\cite{goodfellow2014explaining,amodei2016concrete}. These high-confidence predictions are often produced by the softmax function as it is computed with a fast-growing exponential function. It is clearly necessary to identify potentially wrong predictions. Hendrycks et al.~\cite{hendrycks2017baseline} proposed to detect misclassified samples by enlarging the softmax probabilities between correct and incorrect samples. Meanwhile, utilizing true class probability instead of maximum class probability has been shown to be more reliable in the context of failure detection~\cite{corbiere2019addressing}. In addition, training the model with data that can reflect the complexity of real-world scenario can improve the reliability in prediction, such as curating diabetic retinopathy for training Bayesian DNNs~\cite{band2021benchmarking}. 

To make the model more cautious when it is uncertain, a rejection option allows it to abstain from making a prediction when it is likely to be a mistake. Geifman and El-Yaniv~\cite{geifman2017selective} designed a selective classifier that allows users to set a desired risk level. They further proposed a selective network with a shared classifier of dedicated prediction and ambiguity rejection layer~\cite{geifman2019selectivenet}. What's more, Geifman et.al~\cite{geifmanbias} developed a selective mechanism by using early snapshots for samples with high confidence in model training. 

Besides training classifiers with a rejection option, studies also shed light on post-hoc approaches for failure detection. Setting thresholds based on confidence or uncertainty ranking of samples is widely used to distinguish correct and incorrect predictions, such as AI for breast cancer screening~\cite{leibig2022combining} and decision-making models for low-power Internet of Things (IoT) devices~\cite{cho2020leveraging}. The threshold needs to be tuned as its value trades off the predictor’s coverage rate and the performance on accepted examples~\cite{el2010foundations,hendrickx2021machine}. In our work, we will provide an insightful reference for such threshold selection.


\subsection{Evaluation Metrics}
The quantification of failure detection (FD) performance shares the same characteristic as selective prediction (SP). FD focuses on the model performance after rejecting worst predicted samples under coverage, while SP highlights the model accuracy or error with partial input. More broadly, they are techniques for uncertainty estimation~\citep{ding2020revisiting}. Therefore, the evaluation metrics for SP should also be applicable for FD, such as Area Under the Receiver Operating Characteristic curve (AUROC)~\citep{fawcett2006introduction} and Area Under the Precision-Recall Curve (AUPR)~\citep{manning1999foundations}. Despite the wide use of these metrics for such threshold-independent performance evaluation~\citep{hendrycks2016baseline, malinin2018predictive, leibig2022combining},~\cite{ding2020revisiting} point out that AUROC and AUPR can cause misleading and meaningless results for classification tasks with softmax function. The main reason lies in the assumption that the numbers of correct and wrong predictions are the same. To mitigate this issue, Risk-Coverage (RC) curve is applied for SP in terms of the multi-class classification tasks\citep{geifmanbias, ding2020revisiting, geifman2017selective, zhu2022rethinking}. Hence, this paper utilises the RC curve for the following experiments and analysis.

\subsection{Model Calibration}

To measure the performance of calibration methods, the Expected Calibration Error (ECE)~\cite{naeini2015obtaining} was proposed and is widely applied in various tasks, such as image classification~\cite{geifmanbias,zhu2022rethinking} and sentiment analysis~\cite{muller2019does,obadinma2021class}. ECE splits the data into bins 
, calculates for each bin the average confidence and average accuracy, and averages over all bins.
To alleviate the miscalibration issue for DNNs, calibration techniques have been proposed and then widely applied. Label Smoothing (LS)~\cite{szegedy2016rethinking} reduces over-confidence by computing the cross-entropy loss with uniformly squeezed labels instead of one-hot labels. Extensions of LS such as Margin-based Label Smoothing (MBLS)~\cite{liu2022devil} further provides a unifying constrained-optimization perspective of calibration losses. Focal Loss (FL)~\cite{lin2017focal} adds a focusing factor to the standard cross-entropy loss to deal with an imbalanced dataset. Recent work on sample-dependent focal loss (FLSD)~\cite{mukhoti2020calibrating} investigated the effect of the loss on the training data and achieved impressive performance in calibration. However, it is arguable to what extent calibration techniques can improve the model trustworthiness~\cite{zhu2022rethinking}. Our work will provide a more comprehensive evaluation method regarding this issue. 

\section{Methodology}
\label{method}

The issue we address in this paper is the quantification of the failure detection performance for supervised classification models with the utilization of softmax function. Let $X$ be the input space and $Y=\{1,2,3, \ldots, k\}$ be the set of class labels. Given $D(X, Y)$ as the data distribution over $X \times Y$, a classifier is the function $f$ where the error (true risk) $err$ and accuracy $acc$ is obtained by $f: X \rightarrow Y$. For each input $x \in X$ and its corresponding true label $y$, the probability distribution of the model prediction is $P(y \mid x)$, and the predicted label is $\hat{y}=\operatorname{argmax}_{y \in Y} P(y \mid x)$.

\begin{figure*}[!h]
\centerline{\includegraphics[width=0.95\textwidth]{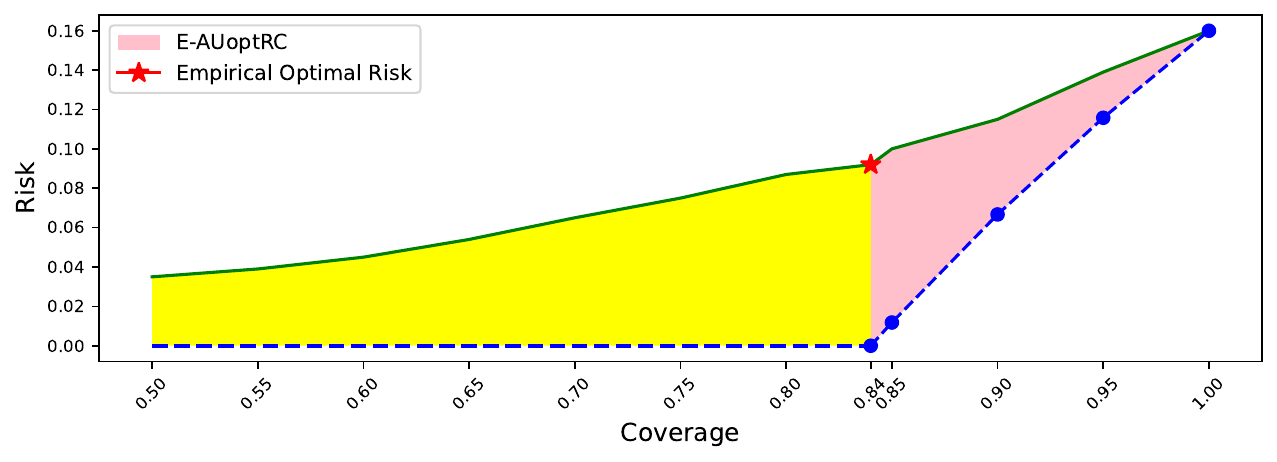}}
\caption{Our proposed method in the RC-curve for ImageNet with SwinTran model. Our proposed E-AUoptRC is shown as the pink area while the E-AURC is the yellow plus pink area. The empirical optimal risk shows the real performance at the optimal point. }
\label{fig:divaurc}
\end{figure*}

\subsection{Problem Setting}

In the Risk-Coverage (RC) curve, the coverage $c$ is the percentage of covered set over the entire data, which is written as $c =\frac{\left|X_c\right|}{|X|}$. For each coverage, the risk is the corresponding error in model prediction. A model with better FD performance should obtain less risk/ higher accuracy with fewer samples rejected. 

To efficiently quantify the FD performance of a model, we first need to construct the reject function $\mathcal{R}$ to decide whether to reject samples or not under different thresholds. By adopting settings in~\citep{lakshminarayanan2017simple, corbiere2019addressing, geifmanbias}, we utilize the predictive uncertainty $u$ to rank samples. A sample with low uncertainty indicates high confidence and better reliability of the model prediction; whereas a sample with high $u$ is more likely to be rejected when narrowing the coverage. Given a fixed or adaptive threshold $t$, the reject function $\mathcal{R}$ is written as follows:

\begin{equation}
\label{equ: classifier}
\mathcal{R}(x) = \begin{cases} \text { cover, }  x \in X_c, & \text { if } u<=t \\
\text { reject, }  x \in X_r, & \text { if } u>t \end{cases}
\end{equation}
where $X_c$ is the covered input set and $X_r$ is the reject set.

There are two types of risks namely empirical risk and optimal risk~\citep{geifmanbias}. The empirical risk $erisk$ is the predicted error of the model under different coverage, as shown in the solid green line in Figure~\ref{fig:aurc}. As the aleatory uncertainty inherits from the data, some risks inevitably exist in certain coverage regardless of the model performance. For a model with perfect uncertainty estimation, if we discard the error percentage of high uncertainty samples, the risk in the remaining coverage input should be zero. This specific coverage point of $1-err$ (or $acc$) was proposed by~\cite{geifmanbias} as the optimal point $op$ and shown as the red star in Figure~\ref{fig:aurc}. Specifically, the risk between coverage of $op$ to 1 monotonically increases until the error of the model. For optimal calibration, the above risks are called optimal risk $optrisk$ illustrated as the blue dotted line in the figure. For example, the model error in the figure is 0.16 and the $op$ is 0.84. Therefore, the optimal risk $optrisk$ under coverage 0 to $op$ is supposed to be 0; while it increases from 0 to 0.16 under $op$ to full coverage. It is worth-noticing that the monotonic increment of $optrisk$ is not exactly in the linear way. 

Both $erisk$ and $optrisk$ can be calculated by Area Under the RC-curve (AURC)~\citep{geifmanbias, ding2020revisiting}, named $empAURC$ (yellow plus grey area in Figure~\ref{fig:aurc}) and $AUOR$ (grey area in Figure~\ref{fig:aurc}) respectively. The difference between $empAURC$ and $AUOR$ is the real FD area, shown as the yellow area in Figure~\ref{fig:aurc}.~\cite{geifmanbias} propose this specific area as the Excess-AURC (E-AURC), where E-AURC $= empAURC - AUOR$.

\begin{table*}[]
\caption{Main Results of AURC, E-AURC, E-AUoptRC, accuracy(ACC) and trust index (TI) on the ImageNet (IN) and Cifar100 (CF100) dataset with CNNs and variants of transformers models. AURC, E-AURC, fE-AURC and lE-AURC are shown as multiply with $10^3$ for clarity.}

\label{main}
\begin{tabular}{l|c|c|c|c|c|c}
\hline
\multicolumn{1}{c|}{\textbf{Dataset}} & \textbf{Model} & \textbf{AURC}                & \textbf{E-AURC}              & \textbf{E-AUoptRC}           & \textbf{ACC(\%)}             & \textbf{TI}                  \\ \hline
IN                                    & DenseNet121    & 93.12                        & 49.13                        & 15.13                        & 71.84                        & 0.856                        \\ \hline
                                      & EfficientNet   & 108.34                       & 75.71                        & 14.81                        & 75.57                        & 0.847                        \\ \hline
                                      & ViT            & {\color[HTML]{FF0000} 40.2}  & {\color[HTML]{FF0000} 25.34} & 6.45                         & 83.26                        & {\color[HTML]{FF0000} 0.906} \\ \hline
                                      & SwinTran       & 53.9                         & 41.03                        & 6.53                         & {\color[HTML]{FF0000} 84.39} & 0.901                        \\ \hline
                                      & CaiT           & 58.29                        & 42.92                        & 6.64                         & 82.99                        & 0.903                        \\ \hline
                                      & CrossViT       & 73.87                        & 56.47                        & 7.79                         & 81.93                        & 0.894                        \\ \hline
                                      & ConvNext       & 56.62                        & 42.13                        & {\color[HTML]{FF0000} 6.38}  & 83.46                        & {\color[HTML]{FF0000} 0.906} \\ \hline
CF100                                 & VGG13\_bn      & {\color[HTML]{FF0000} 75.22} & {\color[HTML]{FF0000} 38.96} & 12.49                        & {\color[HTML]{FF0000} 74.31} & 0.873                        \\ \hline
                                      & VGG19\_bn      & 83.38                        & 45.25                        & {\color[HTML]{FF0000} 11.77} & 73.69                        & {\color[HTML]{FF0000} 0.886} \\ \hline
                                      & ResNet56       & 90.52                        & 47.8                         & 15.02                        & 72.23                        & 0.857                        \\ \hline
                                      & MobileNetV2    & 96.06                        & 48.37                        & 16.41                        & 70.75                        & 0.851                        \\ \hline
\end{tabular}
\end{table*}

\subsection{E-AUoptRC}

The E-AURC reveals the total risk in coverage range from 0 to 1. However, in real-world applications, the coverage is mainly customised due to specific deployment requirements, making it challenging to compare the failure detection (FD) performance for various models. In addition, the E-AURC cannot reveal the failure detection (FD) performance in a specific coverage range. To mitigate the above issues, we propose E-AUoptRC with the coverage from $op$ to 1 (E-AUoptRC, shown as pink in Figure~\ref{fig:divaurc}). We emphasise the E-AUoptRC for the following reasons: (1) it is more practical for deployment, as it is unlikely to discard more than half of data in applications; (2) the smaller E-AUoptRC indicates more samples with high uncertainty are successfully removed so that the model prediction on the remaining data will be more reliable.

\subsection{Trust Index}

Model accuracy $acc$ should track the confidence of the model prediction. For example, a model with 80\% accuracy suggests 80\% confidence in its own predictions, which also defines the perfect confidence score in calibration. As the risk at the optimal point ($ op $) is supposed to be 0, the accuracy at $ op $ should be 1, indicating the prediction's highest model confidence and trustworthiness. In other words, after removing $err\%$ data with high uncertainty, the correctly predicted samples in the remaining data are most trusted. The accuracy at $ op $ also reveals the model calibration, as the discarded $err\%$ data can be misclassified. To represent the model performance in terms of accuracy and calibration, we propose the accuracy at the $ op $ as a Trust Index (TI), a complementary evaluation to the accuracy metric to indicate the model's trustworthiness. For example, in Figure~\ref{fig:divaurc}, with the model accuracy of 84\%, the model is 0.84  trust of the prediction. After removing 16\% samples with high uncertainty (the $ op $ is 0.84), the risk is approximately 0.08. The $TI$ , the accuracy over the most confident 84\% of samples is 0.92. The higher TI suggests the better trustworthiness of the model predictions, and we next present empirical data to substantiate this.

\section{Experimental Setup}

\subsection{Datasets and Baselines}

We validate the proposed method with three benchmark image datasets: ImageNet 2012 (IN)~\cite{russakovsky2015imagenet}, CIFAR100 (C100)~\cite{krizhevsky2009learning} and Tiny-ImageNet ~\cite{deng2009imagenet}. For baselines, we use state-of-the-art (SOTA) Vision Transformer (ViT)~\cite{dosovitskiy2020image} and its variants such as Swin-Transformer (SwinT)~\cite{liu2021swin}, Class-Attention in Image Transformers (CaiT)~\cite{touvron2021going}, Cross-Attention Multi-Scale Vision Transformer (CrossViT)~\cite{chen2021crossvit}, ConvNext~\cite{liu2022convnet} with the ImageNet pretrained weights from TIMM~\footnote{https://github.com/rwightman/pytorch-image-models} library. To report comprehensive results on various models architectures, we also use the Convolutional neural networks (CNNs) in our experiments, namely DenseNet121~\cite{huang2017densely}, ResNet56~\cite{he2016deep}, variants of VGG~\cite{simonyan2014very} and MobileNetV2~\cite{sandler2018mobilenetv2}. All models are with pretrained weights of ImageNet dataset. For recent SOTA calibration techniques label smoothing (LS)~\cite{szegedy2016rethinking}, focal loss (FL)~\cite{lin2017focal}, MBLS~\cite{liu2022devil} and FLSD~\cite{mukhoti2020calibrating}, we utilize the pre-trained model and official implementation from the repository~\footnote{https://github.com/by-liu/MbLS}. 

As the evaluation of failure detection is a post-processing approach, we primarily utilize each dataset's test set. For the ImageNet dataset, we equally divide its original test set of 50,000 images into validation and test sets for a fair comparison. For Tiny-ImageNet and CIFAR100 dataset, an 80/10/10 for training/validation/test split is applied. 

\subsection{Implementation Details}

For a fair comparison and replicability of experimentation, we utilized publicly available existing pre-trained weights for our investigation and experimentation. The GPU of the Nvidia Tesla P40 was used for all experiments. The bins number for ECE was set as $M=15$.

\begin{table*}[]
\caption{Results for SOTA calibration techniques on failure detection with Tiny\_ImageNet dataset with ResNet50 model. AURC, E-AURC, fE-AURC and lE-AURC are shown as multiply with $10^3$ for clarity. ECE\_OP denots the ECE at the optimal point. ECE and ECE\_OP are shown in percentage.}
\label{compare}
\begin{tabular}{c|c|c|c|c|c|c|c}
\hline
\textbf{Method} & \textbf{AURC}                 & \textbf{E-AURC}              & \textbf{E-AUoptRC}           & \textbf{ACC(\%)}             & \textbf{TI}                  & \textbf{ECE(\%)}            & \textbf{ECE\_OP(\%)}        \\ \hline
CE              & {\color[HTML]{FF0000} 128.71} & {\color[HTML]{FF0000} 57.94} & 22.13 & 64.82                        & 0.821 & 3.76                        & 4.25                        \\ \hline
LS              & 131.54                        & 63.51                        & {\color[HTML]{FF0000} 21.98}                        & {\color[HTML]{FF0000} 65.46} & {\color[HTML]{FF0000} 0.824}                        & 2.8                         & 2.04                        \\ \hline
MBLS            & 135.39                        & 64.27                        & 22.78                        & 64.74                        & 0.817                        & {\color[HTML]{FF0000} 1.87} & {\color[HTML]{FF0000} 0.92} \\ \hline
FL              & 146.42                        & 68.61                        & 25.05                        & 63.24                        & 0.807                        & 3.1                         & 3.53                        \\ \hline
FLSD            & 139.72                        & 64.85                        & 23.91                        & 63.89                        & 0.812                        & 2.8                         & 2.49                        \\ \hline
\end{tabular}
\end{table*}

\begin{figure*}[!h]
\centerline{\includegraphics[width=1\textwidth]{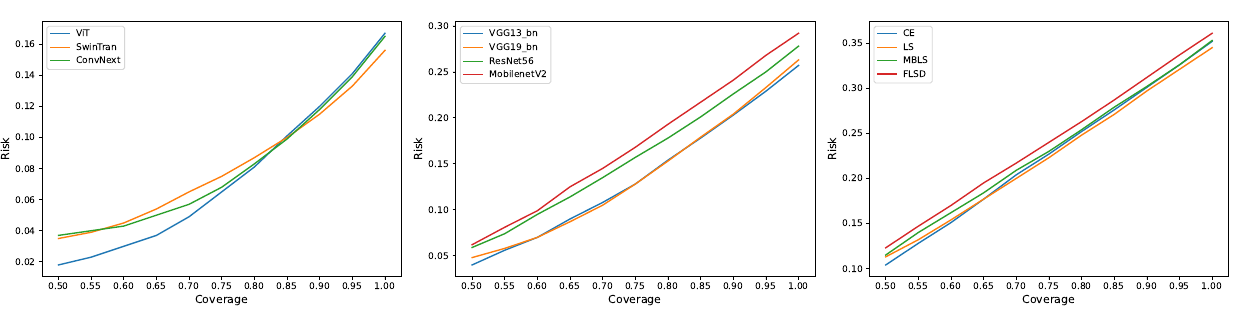}}
\caption{Risk-coverage curve for the visualization of failure detection performance. Left: ImageNet dataset with transformer models of ViT, SwinTran and ConvNext; Middle: Cifar100 dataset with CNNs models of VGG13\_bn, VGG19\_bn, ResNet56 and MobileNetV2. Right: Tiny\_ImageNet with ResNet50 model on SOTA calibration techniques. CE, LS, MBLS, FLSD denots baseline, lable smoothing, margin-based label smoothing and sample-dependent focal loss respectively. The coverage starts from 0.5 instead of 0 for the clarity of visualization. }
\label{fig:rc_all}
\end{figure*}

\section{Results}

We conducted extensive experiments on benchmark datasets ImageNet and Cifar100 with various CNNs and variants of transformers to compare the AURC, E-AURC and our proposed E-AUoptRC. We further observed the limitation of the conventional overall model accuracy and how our proposed Trust Index (TI) mitigates it. Finally, to validate the efficacy of our method, we applied it to SOTA calibration techniques with Tiny\_ImageNet on ResNet50 dataset. All the experiments and results are shown in Tables ~\ref{main} and~\ref{compare}, and Figure~\ref{fig:rc_all}.

Table~\ref{main} shows the results for image classification with the benchmark datasets. For AURC, E-AURC and E-AUoptRC in the ImageNet dataset, the variants of transformers outperform CNNs model. 
The E-AURC for ViT is about half of the E-AURC of SwinTran, CaiT and ConvNext, indicating that ViT greatly outperforms the other three models in failure detection. However, regarding the E-AUoptRC, the difference is almost ignorable and the ConvNext is slightly better than the other three models. The risk-coverage (RC) curve (Left in Figure~\ref{fig:rc_all}) also shows that at the coverage of 0.84 (near the optimal point) to 1, the risk curve of ViT and ConvNext is nearly overlapping. The lower risk for VIT occurs at very low coverage levels, which are not of interest for most real world applications. For CF100 dataset with CNNs, VGG13\_bn substantially outperforms other models in terms of AURC and E-AURC. However, the difference in E-AUoptRC between VGG13\_bn and VGG19\_bn is much smaller. This can be understood from the Middle plot in Figure~\ref{fig:rc_all}, where the curve for VGG13\_bn and VGG19\_bn overlaps at coverage between 0.74(near the optimal point) to 0.9. These differences in the metrics provide empirical evidence  that our proposed E-AUoptRC more accurately reflects real differences in failure detection performance than other methods. 

Similar to the results of AURC-related evaluation, the variants of transformer models also outperform CNNs in terms of overall model accuracy and trust index (TI). The SwinTran obtains the highest overall model accuracy for the ImageNet dataset, but it does not yield the highest TI. For the Cifar100 dataset, the VGG13\_bn achieves the highest overall model accuracy, whereas the VGG19\_bn obtains the best TI. It indicates that the model with the highest overall accuracy does not guarantee the highest TI, which shows that our proposed TI is necessary for model trustworthiness evaluation. 

In Table~\ref{compare}, the baseline (CE) obtains better AURC and E-AURC, but label smoothing outperforms other methods and CE in terms of overall accuracy (improves by 0.6\%) and TI. MBLS nearly halves the overall ECE of baseline and achieves the best ECE at the optimal point. In the Right RC curve in Figure~\ref{fig:rc_all}, LS is with the lowest risk at the coverage of 0.65 to 1 (the likely operating range when the model is deployed), and our proposed E-AUoptRC and TI metrics are the only ones that capture this. Failure detection performance should be a significant evaluation for calibration techniques, and our methods provide a more insightful view of the model trustworthiness. 

\section{Discussion \& Conclusion}

In this paper, we proposed the E-AUoptRC to more precisely quantify the failure detection performance in the key region of interest, and the Trust Index (TI) that measures model accuracy at its optimal point. The empirical results show that our methods can better reveal the model trustworthiness under a fair comparison. In the real-world deployment, a fixed threshold is often used due to specific task requirements and simplicity of implementation. Our proposed TI can be utilized as the reference for the threshold selection with following reasons: (1) the accuracy should indicate the model confidence in its prediction, suggesting the TI can interpret the confidence; (2) TI is obtained at the optimal point, where the model is supposed to achieve the ideal performance. This is an objective method for the fair comparison of models with different accuracy and calibration (as shown in Table~\ref{main} and~\ref{compare}); (3) TI is easy to calculate, which is a time and computational cost saving. We have shown several benefits of our proposed metrics over existing ones and in our future work, we will further investigate the role of TI in improving failure detection.

\bibliography{references}

\appendix

\end{document}